\documentclass{article}

\usepackage{arxiv}

\usepackage[utf8]{inputenc} 
\usepackage[T1]{fontenc}    
\usepackage{hyperref}       
\usepackage{url}            
\usepackage{booktabs}       
\usepackage{amsfonts}       
\usepackage{nicefrac}       
\usepackage{microtype}      
\usepackage{lipsum}
\usepackage{graphicx}
\graphicspath{ {./images/} }

\title{A Survey on Differential Privacy with Machine Learning and Future Outlook}

\author{
 Samah Saeed Baraheem \\
  Department of Computer Science\\
  University of Dayton\\
  Dayton, OH 45469 \\
  \texttt{baraheems1@udayton.edu} \\
   \And
 Zhongmei Yao \\
  Department of Computer Science\\
  University of Dayton\\
  Dayton, OH 45469 \\
  \texttt{zyao01@udayton.edu} \\
}

\begin{document}
\maketitle
\begin{abstract}
Nowadays, machine learning models and applications have become increasingly pervasive. With this rapid increase in the development and employment of machine learning models, a concern regarding privacy has risen. Thus, there is a legitimate need to protect the data from leaking and from any attacks. One of the strongest and most prevalent privacy models that can be used to protect machine learning models from any attacks and vulnerabilities is differential privacy (DP). DP is strict and rigid definition of privacy, where it can guarantee that an adversary is not capable to reliably predict if a specific participant is included in the dataset or not. It works by injecting a noise to the data whether to the inputs, the outputs, the ground truth labels, the objective functions, or even to the gradients to alleviate the privacy issue and protect the data. To this end, this survey paper presents different differentially private machine learning algorithms categorized into two main categories (traditional machine learning models vs. deep learning models). Moreover, future research directions for differential privacy with machine learning algorithms are outlined. 
\end{abstract}


\section{Introduction}
Machine learning has proven its capability in effectively solving real-world problems, even with previously unsolvable problems. The objective of machine learning is to simulate and imitate the human behaviors so that machines and computers are able to learn and acquire new skills and/or knowledge from the given data. Therefore, machine learning has been, and still is, an active and hot topic among researchers. Recently, more and more machine learning models and applications have been developed and deployed. However, vulnerabilities and privacy leaks might be a serious threat to the participants' data. Particularly, when the dataset contains sensitive personal information. For instance, for health care applications, the dataset might include some very sensitive information, such as patient names, contact phone numbers, addresses, email addresses, dates of birth, insurance details, photo ID, and tax ID numbers. Indeed, machine learning task has the ability to extract useful and meaningful information from data to learn new knowledge/skills; and thus, it enhances the progress in companies, commerce, industry, academia, and science. Nonetheless, the ability of machine learning to capture fine-grained details may compromise data providers' privacy.
\\Several research \cite{Fredrikson_Jha_Ristenpart_2015,DBLP:journals/corr/ShokriSS16,Hitaj_Ateniese_Perez-Cruz_2017} states that inferring data about specific records in the training dataset is possible even with black-box settings. Different types of attacks might threaten data privacy. These attacks can be categorized into two categories: passive adversaries, i.e., model inversion \cite{Fredrikson_Jha_Ristenpart_2015} and membership inference \cite{DBLP:journals/corr/ShokriSS16}, and active adversaries, such as \cite{Hitaj_Ateniese_Perez-Cruz_2017}, just to name a few. Fredrikson et al. \cite{Fredrikson_Jha_Ristenpart_2015} introduced a model inversion attack. This attack can reveal the faces of participants by providing the face recognition system API along with the participant's name. Shokri et al. \cite{DBLP:journals/corr/ShokriSS16} proposed a membership inference attack. The attack is capable of predicting if the training dataset includes a particular record though black-box access to a machine learning model. Hitaj et al. \cite{Hitaj_Ateniese_Perez-Cruz_2017} proposed a vigorous attack against collaborative/distributed deep learning using GANs that results in inferring sensitive information from the user's device.
\\Hence, there is an urgent demand for privacy protection and preservation. Various solutions and mechanisms have been proposed to tackle this serious problem. One of the strongest and most popular solutions is differential privacy (DP) \cite{Dwork_2006}. DP is a powerful technique for privacy guarantees. It was initially used in simple statistics \cite{Dwork_Lei_2009,Dwork_Smith_2010}, but then the research community has leveraged DP in a machine learning environment \cite{Abadi_Chu_Goodfellow_McMahan_Mironov_Talwar_Zhang_2016,Papernot_Abadi_Erlingsson_Goodfellow_Talwar_2016,Papernot_Song_Mironov_Raghunathan_Talwar_Erlingsson_2018,McMahan_Ramage_Talwar_Zhang_2017} to protect the dataset from being revealed 

\begin{figure} 
    \centering
    \includegraphics[scale=0.6]{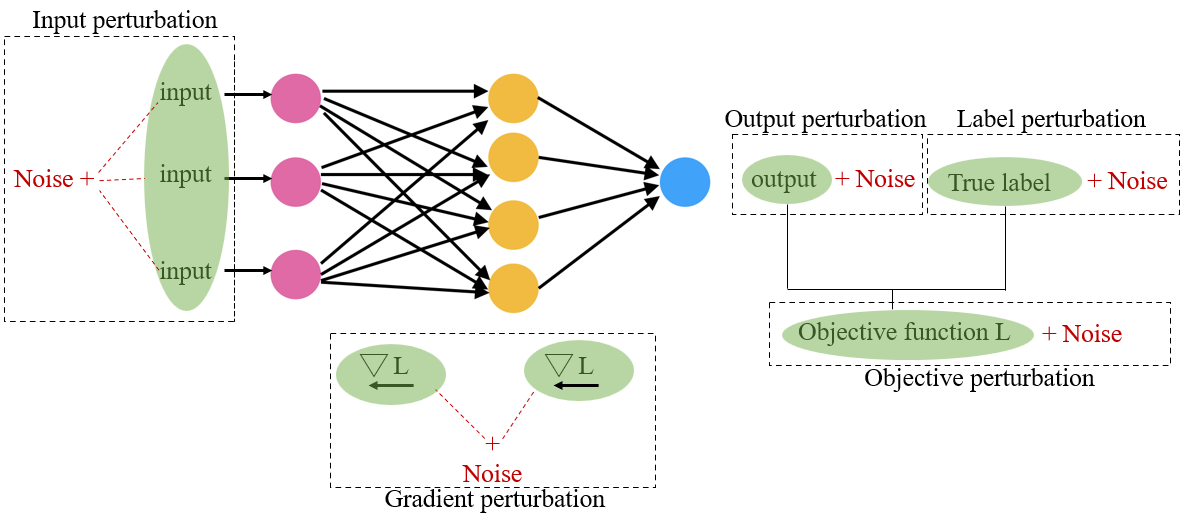}
    \caption{An overview of the differential privacy positions in the machine learning pipeline.}
    \label{fig:fig1}
\end{figure}
\begin{figure} 
    \centering
    \includegraphics[scale=0.6]{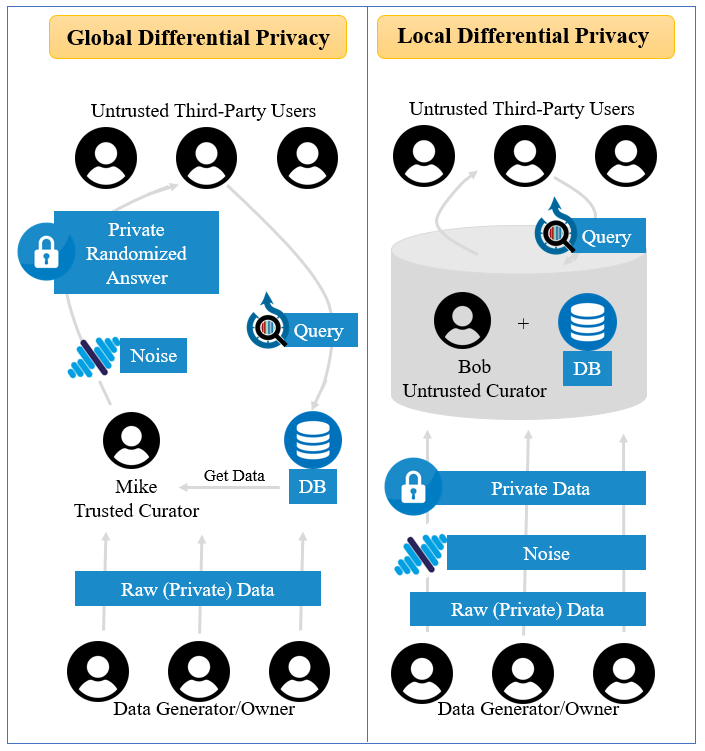}
    \caption{The difference between global differential privacy (GDP) and local differential privacy (LDP).}
    \label{fig:fig2}
\end{figure}


based on the outcomes. DP works by injecting an additional statistical noise to the dataset, whether to the inputs, the outputs, the ground truth labels, the objective functions, or the gradients, as shown in Figure \ref{fig:fig1}. \\Furthermore, adding noise could be accomplished locally (client side) or globally (server side). Based on this, DP is categorized into two categories, global differential privacy (GDP) and local differential privacy (LDP), as illustrated in Figure \ref{fig:fig2}. GDP requires a trusted data curator. The perturbation occurs at output time in the server side and by the data curator. This leads to more accurate results because the amount of added noise is not significant and happens at the end of the process. However, it is more vulnerable to attacks because the data curator needs the original data to add statistical noise. Thus, this might make the system vulnerable to attacks, i.e., membership inference \cite{DBLP:journals/corr/ShokriSS16} and model memorizing attacks \cite{Leino_Fredrikson_2020}. As a result, this makes GDP mechanisms unsuitable for machine learning applications \cite{DBLP:journals/corr/abs-1908-02997}. On the other hand, in LDP, there is no need to have data curator since the perturbation step is occurred at input time in client side and by the data owner. Thus, it is a more preservative and secure model. Therefore, it is more suitable for machine learning algorithms
\\The remaining sections of this paper are as follows. We first review various differentially private machine learning algorithms. Then, we conclude by suggesting possible future research directions for differential privacy with machine learning algorithms.

\section{Differentially Private Machine Learning Algorithms}
\label{sec:others}
\begin{figure} 
    \centering
    \includegraphics[scale=0.6]{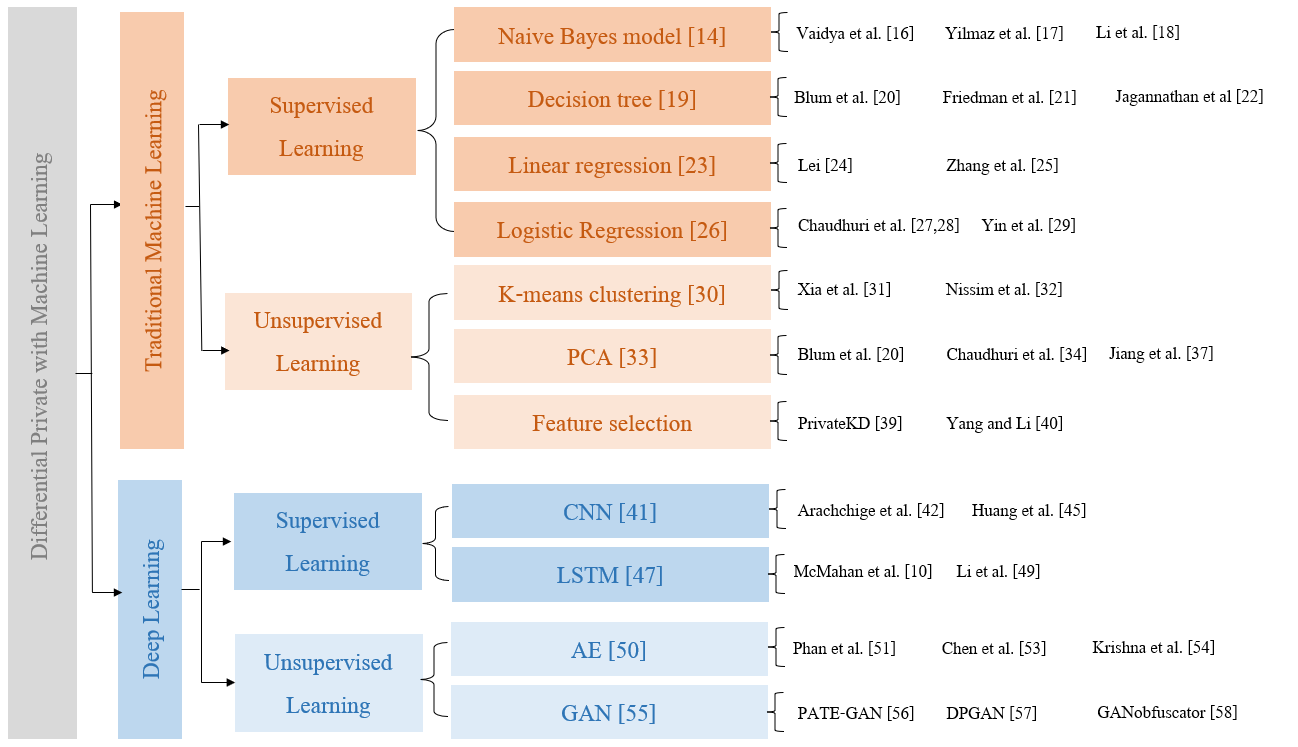}
    \caption{A summary of differentially private machine learning algorithms' taxonomy.}
    \label{fig:fig3}
\end{figure}

Machine learning is a powerful technique that extracts useful information about the underlying distribution from the given dataset. Recently, machine learning has attracted the researchers’ attention due to its capability to solve even the previous unsolvable problems. However, machine learning models might leak the training data. This could be a serious threat if the information contained in the training dataset is sensitive and private. Therefore, many research has been recently conducted to protect the training data. One of the popular solutions to preserve the dataset from leaking while maintain the dataset quality is differential privacy (DP). Therefore, in this section, various differential privacy with machine learning algorithms are introduced. In general, DP guarantees privacy in machine learning models through adding noise to the inputs, outputs, ground truth labels, or even to the models. Thus, adversaries are unable to predict and infer any information for any individual record after publicly releasing the machine learning models or only the results. Figure \ref{fig:fig3} summarizes the taxonomy of differentially private machine learning algorithms. In this survey paper, differentially private machine learning algorithms are categorized into two groups based on the machine learning types (traditional or deep learning).

\subsection{Differential Private Traditional Machine Learning}
\subsubsection{Differential Private Supervised Learning}
Classification and regression models infer a variable Y based on a given combination of variables X. While the former infers categorical variable, the latter estimates a numerical variable. 
\paragraph{Naive Bayes model \cite{Lowd_Domingos_2005}.} It is one of the traditional machine learning algorithms which used for classification purpose. Specifically, it is a family or a collection of simple probabilistic algorithms based upon Bayes' theorem \cite{ Coletti_Scozzafava_2002}. Bayes' theorem is based on computing conditional probabilities, where it attempts to find the probability of an event that is happening provided the probability of another event that has previously happened. Hence, it computes the conditional probability for all possible labels. The label with the highest probability is the predicted label.
\\To maintain the privacy, $\epsilon$-differential privacy is leveraged with Naive Bayes classifier \textbf{\cite{Vaidya_Shafiq_Basu_Hong_2013}}. The idea behind it is based on the bound assumption, meaning that all features values in the training set are bounded by specific value. It could be based on the Gaussian assumption as well if the bound contains the Gaussian distribution. Next, the information sensitivity is computed, and Laplace noise is added to this information to protect the model from being compromised. Moreover, to preserve the individuals’ training data, \textbf{Yilmaz et al. \cite{Yilmaz_Al-Rubaie_Chang_2019}} propose a local differential privacy (LDP) with Naive Bayes algorithm using locally differentially private frequency and statistics estimation mechanisms. This approach helps in collecting the training data to be used in Naive Bayes classifier while preserving privacy of individuals who provide the training data. In this proposed method, perturbed data from participants are first collected, where the relationships between the feature values and the corresponding labels should be maintained during collecting data. In order to maintain this relationship, each feature value and label from the user are transformed first into a new value. Then, LDP perturbation is performed. Five LDP techniques are used for perturbation which are Direct Encoding (DE), Symmetric and Optimal Unary Encoding (SUE and OUE), Summation with Histogram Encoding (SHE), and Thresholding with Histogram Encoding (THE). It works well with discrete and continuous data. However, for continuous data, the authors further apply dimensionality reduction techniques to enhance the accuracy. 
\\As opposed to \cite{Vaidya_Shafiq_Basu_Hong_2013,Yilmaz_Al-Rubaie_Chang_2019}, where the data are privately collected from a single data provider, \textbf{Li et al. \cite{Li_Li_Liu_Li_Jia_2018}} propose a DP with Naive Bayes in multi-data provider setting. It works by initializing the cryptographic tools along with the public parameters. After that, each data provider encrypts their data and sends the encrypted data to the data collector. The data collector aggregates all received data from multiple owners and adds the Laplace noise based on the auxiliary information. This leads to achieve the privacy-preserving solution. Following this, the trainer receives the joint dataset from the data collector and trains the model over different aggregated dataset while preserving the ownership privacy. Thus, adversaries are unable to infer whether a specific data owner holds a record that contains a particular feature value.

\paragraph{Decision tree \cite{Myles_Feudale_Liu_Woody_Brown_2004}.} It is a special form of probability tree which used to make a decision or a prediction based on prior knowledge. In particular, the data are split recursively based on the chosen features; and thus, the learning is an iterative and repetitive process. It first starts from the root which contains the whole training dataset. Then, based on the selected feature, the training set is segmented into sub-sets, where each and every sub-set satisfies the best classification. This process of choosing a feature and splitting the set into sub-sets is iterative. After that, a pruning process begins to eliminate unnecessary sub-sets from the tree and merge them to their parent nodes.
\\The first decision tree model without compromising the privacy is proposed in \textbf{\cite{Blum_Dwork_McSherry_Nissim_2005}}.
This model is based on the Sub-Linear Queries (SuLQ) framework that protect the privacy through adding noise to the features' information gain.
To split a node into sub-nodes, the model selects a feature with less value than a particular threshold.
Since the noise is added iteratively to the features in each splitting step, it may fail due to the large volume of added noise.
To tackle this problem, \textbf{Friedman et al. \cite{Friedman_Schuster_2010}} leverage an exponential technique in the feature selection step.
This leads to less amount of noise than \cite{Blum_Dwork_McSherry_Nissim_2005}. However, \cite{Friedman_Schuster_2010} still results in a large amount of noise. Thus, to reduce the amount of noise in the feature selection step, \textbf{\cite{Jagannathan_Pillaipakkamnatt_Wright_2012}} is proposed. In random decision tree with DP, the pruning phase is removed by eliminating the empty nodes. Then, reconstructing the tree in a way that all leaf nodes are put at the very same level. Following this, a Laplace noise is added to each leaf nodes which satisfies $\epsilon$-DP.

\paragraph{Linear regression \cite{Su_Yan_Tsai_2012}.} It is a simple traditional machine learning algorithm that is mainly used for predicting a numerical feature value (dependent variable) based on values of other features (independent variable(s)). If the number of independent variable is one, then it is called a simple linear regression. Meanwhile, when the number of independent variable is more than one, it is called multiple linear regression. As a result, both types of linear regression work by computing the conditional probability distribution of dependent variable given the values of the independent variable(s).
\\In \textbf{\cite{ Lei_2011}}, the privacy of linear regression models is preserved by generating a synthetic training dataset. Instead of using the original dataset, the data is augmented with noise. In particular, the noise is added to the inputs' histograms. This perturbed histogram guarantees privacy, but it only works well with low-dimensional training set. Additionally, it sometimes fails and obtains inaccurate regression outputs as if there is no privacy-preserving mechanism.
For the above aforementioned reason, \textbf{\cite{ Zhang_Zhang_Xiao_Yang_Winslett_2012}} was proposed based on functional mechanism. In this work, the objective/cost function is perturbed by inserting noise to its coefficients after approximating the objective function through Taylor expansion to approximate it to a low order. Then, it attempts to find the weight which minimizes the approximated objective function. This leads to $\epsilon$-differentially private linear regression model.

\paragraph{Logistic Regression \cite{LaValley_2008}.} Likewise linear regression, logistic regression works by predicting the value of the dependent variable based on its relationship to one or more independent variables. The difference is that it only works to infer a categorical variable rather than a continuous variable. Mostly, in logistic regression, a regularization term is incorporated into the adjective function to overcome the problem of overfitting. Hence, applying output perturbation or objective perturbation which add noise to the output or adjective/cost function, respectively as \textbf{in \cite{ NIPS2008_8065d07d, Chaudhuri_Monteleoni_Sarwate_2009}} can ensure privacy preserving over regularized logistic regression. However, this mechanism will not work on standard logistic regression. To tackle this problem, \textbf{Yin et al. \cite{ Yin_Zhou_Yin_Wang_2019}} propose to use local differential privacy (LDP) to protect the privacy of regularized and standard logistic regression. Three steps are employed, starting with adding noise, followed by selecting features, and ending with training the logistic regression model. In the first phase, Laplace mechanism is used to add noise to the training dataset. Then, feature selection starts to guarantee that the noisy data is not removed and to enhance the classification performance. At the end, a logistic regression classifier is trained on the perturbed data.
 
\subsubsection{Differential Private Unsupervised Learning}
Unsupervised traditional machine learning algorithms can be divided into three main categories, clustering, dimension reduction, and feature selection. Clustering models cluster and group unlabeled data into different clusters/groups depending on the similarity. Whereas dimension reduction models attempt to project the data from a high-dimensional space to a low-dimensional space by preserving the most useful and meaningful data, feature selection mechanisms attempt to select the most informative variables/features from the given dataset. Thus, both dimension reduction and feature selection are used as a pre-processing step for further studies.
\paragraph{K-means clustering \cite{Alsabti_Ranka_Singh_1997}.} It is an unsupervised learning method that works on unlabeled datasets. Thus, it works without intervention of human/annotator. Specifically, it iteratively groups the datapoints into different clusters based on the similarity, where the number of clusters (k) is predefined. Each datapoint belongs to one cluster based on the nearest centroid. Then, at each iteration, each centroid is updated based on the mean of datapoints in that cluster. 
\\\textbf{Xia et al.\cite{Xia_Hua_Tong_Zhong_2020}} propose the first LDP K-means clustering. They leverage LDP for privacy preservation of K-means clustering through direct and local perturbation mechanism over the training dataset. Furthermore, a budget allocation technique is used to improve the accuracy by decreasing the noise scale. An extended version is introduced in the same work to enhance both the utility and privacy. In the improvement version, not only the training data is perturbed, but also the intermediate outcomes of clusters in each iteration are perturbed. \textbf{Nissim et al. \cite{Nissim_Raskhodnikova_Smith_2007}} introduce differential privacy with K-means clustering through smooth sensitivity and sample-aggregate frameworks. Particularly, sample-aggregate framework randomly divides the training set into several subsets, and then apply K-means clustering on each subset individually. Thus, it produces several outputs, one output for each subset. Following this, a smooth sensitivity framework is used to add instance-specific noise to each output. This leads to add a control on the amount of noise; and thus, it improves the accuracy. Additionally, it aids in publishing the output from a differential private dense set which leads to preserve the privacy while maintaining the K-means clustering.
\paragraph{Principal component analysis (PCA) \cite{Abdi_Williams_2010}} is a popular dimension reduction model that is used to transform from a high-dimension to a low-dimension while maintaining most of the information in the original dataset. Therefore, it helps in analyzing and easily visualizing the dataset. 
\\To preserve privacy while maintaining the performance of PCA, \textbf{Blum et al. \cite{Blum_Dwork_McSherry_Nissim_2005}} propose to add noise to the second moment matrix. Following this, PCA is run on the perturbed matrix. However, because of the noise amount added, this approach might significantly impact the approximation quality. To tackle this issue, \textbf{Chaudhuri et al. \cite{Chaudhuri_Sarwate_Sinha_2012}} present PPCA. This method is based on the exponential mechanism in \cite{Austrin_2007} which maintain the data privacy. In PPCA, it randomly samples a k-dimensional subspace from the matrix Bingham distribution \cite{Chikuse_2003}. This distribution not only ensures privacy, but also ensures the quality of approximation. Hence, PPCA is a differential privacy PCA method. To enhance the privacy guarantee, \textbf{\cite{Jiang_Xie_Zhang_2016}} is introduced. In this approach, a novel input perturbation technique is proposed for obtaining a covariance matrix that achieves ($\epsilon$,0)-differential private. Specifically, the Wishart distribution \cite{ Olkin_Rubin_1962} is utilized to produce noise. Then, the Wishart noise matrix is added to the original covariance matrix, providing a noisy covariance matrix prior finding the eigenspace.
\paragraph{Feature selection.} Feature selection is the process of reducing the number of inputs that are fed into the model later. In particular, it is the method for the automatic selection of the most informative and relevant features.
\\Vinterbo proposes \textbf{PrivateKD \cite{Vinterbo_2012}}, a differential privacy feature selection for classification. Two assumptions are held in PrivateKD. The first assumption is that all variables are categorical. The second assumption is that each variable is limited to certain potential values. The idea behind private projected histogram (PPH) is that it first specifies the number of features to be selected. Then, it incrementally selects the features. Specifically, the selected features set is first initialized to empty set, and then it adds new features one at a time through a greedy method along with the exponential mechanism to increase the distinguishability of the selected features. Thus, the new perturbed dataset is generated before feeding it to a classifier. However, this method might not work well with high-dimensional features. \textbf{Yang and Li \cite{Yang_Li_2014}} propose a differential privacy feature selection algorithm. It relies on local learning, where each datapoint is grouped to the nearest neighbor using the Manhattan distance. This results in scaling each feature; and thus, it produces a weighted feature space. Moreover, to fit the model and overcome the overfitting problem, the logistic regression loss with L2-regularizer is incorporated. To guarantee privacy, output perturbation mechanism is used to add noise to the output based on the sensitivity analysis of the model.

\subsection{Differential Private Deep Learning}
Recently, due to its effectiveness and powerfulness when trained on a large dataset, deep learning has attracted many researchers in image classification, object detection, and natural language processing; just to name a few. Nevertheless, the training sensitive data are at risk of adversaries. Therefore, to alleviate privacy concerns, differential privacy mechanisms are leveraged with deep learning. In this section, different differential private deep learning models are briefly summarized into two categories (supervised and unsupervised models).
\subsubsection{Differential Private Supervised Learning}
\paragraph{Convolutional Neural Network (CNN) \cite{Lecun_Bottou_Bengio_Haffner_1998}.} It is a special type of artificial neural network (ANN). It is commonly used to analyze the visual data in many applications, such as classification, recognition, and so on. The reason of its popularity is that it has the ability to automatically extract the important features during the training process.
\\To ensure privacy-preserving deep learning, i.e., CNN, \textbf{Arachchige et al. \cite{Arachchige_Bertok_Khalil_Liu_Camtepe_Atiquzzaman_2020}} present a local differential privacy (LDP) deep learning algorithm named LATENT. This method allows the data providers to insert a randomization layer (i.e., LDP layer) prior data leave their devices and before the data are sent to a machine learning service. Thus, the data providers perturb the data prior releasing them. This way the data are preserved from leaking in the server side. It efficiently works with a convolutional neural network (CNN), where it first divides the CNN architecture into three layers. The first layer is the convolutional module. The second layer is the randomization module (i.e., LDP layer) which adds a privacy preserving service. The third and last layer in CNN architecture is the fully connected module. The second layer leverages the features of randomized response \cite{Fox_2015} which guarantees LDP and uses a new LDP protocol called utility enhancing randomization (UER). UER depends on a modified version of optimized unary encoding protocol (OUE) \cite{Wang_Blocki_Li_Jha_2017} to enhance the flexibility in selecting randomization probabilities by adding a coefficient epsilon (privacy budget). \textbf{Huang et al. \cite{Huang_Guan_Zhang_Qi_Wang_Liao_2019}} propose a new optimization algorithm for CNN, named DPAGD-CNN. DPAGD-CNN is short for Differential Privacy Adaptive Gradient Descent for Convolutional Neural Network. Instead of using a fixed privacy budget in each iteration during training process, DPAGD-CNN adaptively allocates different privacy budget per iteration. The privacy budget is split into two parts in each iteration. While one part is used to calculate the noisy gradient, the rest is used to choose the optimal step size. In particular, in each step, different noise is injected into the gradient via adaptive approach, but the overall privacy budget should be the same. The amount of noise is bigger than gradient norm in the first iteration of the optimization since it will not impact the gradient descent direction at the beginning. However, the noise is reduced in latter iterations to ensure precise gradient descent direction. Zero-concentrated differential privacy (ZCDP) \cite{Bun_Steinke_2016}, a relaxed variant of differential privacy, is used.
\paragraph{Long Short-Term Memory Networks (LSTM) \cite{Hochreiter_Schmidhuber_1997}.} It is a type of recurrent neural network (RNN) that is able to learn long-term dependencies. Thus, LSTM is able to memorize the past and then find out patterns throughout time to obtain next guesses that make sense. It is widely used in machine translation, speech recognition, language modeling, and sentiment analysis, just to name a few. 
\\\textbf{McMahan et al.} propose differentially private LSTM language models \textbf{\cite{McMahan_Ramage_Talwar_Zhang_2017}}. This model preserves privacy while it maintains the accuracy using LSTM to predict the next word in mobile keyboards. User-level privacy guarantees are achieved based on the federated averaging algorithm \cite{McMahan_Moore_Ramage_Arcas_2016}. The federated averaging algorithm gathers many stochastic gradient descent (SGD) updates which in turns are averaged to calculate the final update.  The final update is perturbed using Gaussian noise. Moreover, each-user update is clipped to make the total update bounded in L2 norm. The federated average algorithm leads to large-step updates; and hence, fewer training steps which results in better accuracy and privacy. As a result, this model achieves differential privacy without reducing the accuracy. However, the computation is increased, leading to increase the training time. \textbf{Li et al. \cite{Li_Li_Yang_Yang_Liu_2019}} present differential privacy deep learning called (DP-LSTM) for predicting stock price based on financial news articles and sentiment analysis. In this paper, the first step is to formulate a sentiment-ARMA based on the autoregressive moving average model (ARMA) that considers the financial news articles information to extract the information and analyze the sentiment. Following this, an LSTM is implemented. This LSTM model includes three components which are VADER, LSTM, and DP mechanism. The model uses valence aware dictionary and sentiment reasoner (VADER) to compute the sentiment scores. VADER is a rule-based sentiment analysis and lexicon. To improve the robustness of LSTM predictions and guarantee privacy, DP mechanism is adopted by injecting noise from Laplace distribution. The model works well in predicting the stock price with less errors and high robustness.

\subsubsection{Differential Private Unsupervised Learning}
\paragraph{Autoencoders (AE) \cite{ Bank_Koenigstein_Giryes_2020}.} It is a type of unsupervised learning, where the training dataset only contains inputs without outputs/labels. The target values (outputs) are set to be equal to the corresponding inputs, leading to forming the task as a supervised learning task. Then, the model attempts to minimize the reconstruction error which is the difference between the original inputs and the corresponding reconstructed outputs. Therefore, it attempts to learn compressed representations of the original inputs. In general, it has two components (an encoder and a decoder). The encoder attempts to transform a high-dimensional sample to a low-dimensional representation, while the decoder attempts to reconstruct the high-dimensional sample from the low-dimensional representation. 
\\\textbf{Phan et al. \cite{Phan_Wang_Wu_Dou_2016}} propose a DP-based deep autoencoders named deep private autoencoder (dPA) to predict the human behavior in a health social network while preserve privacy.  Deep autoencoders \cite{Bengio_2009} is a model that consists of many autoencoders. The objective is to extract useful and meaningful latent representations through an unsupervised learning. To add a privacy level to deep autoencoders in order to protect the data, objective perturbation is leveraged. Thus, it first approximates the cross-entropy objective function of the reconstructed inputs to a polynomial approximation via Taylor series. After that, noise is added into the coefficients of the polynomial approximation objective function. Finally, a normalization layer is added into the private autoencoder (PA) to guarantee privacy in a deep autoencoders since multiple private autoencoders can be stacked on top of each other. \textbf{Chen et al. \cite{ Chen_Xiang_Xue_Li_Borisov_Kaarfar_Zhu_2018}} propose a differentially private autoencoder-based generative model (DP-AuGM). In this model, a private data using a differentially private algorithm \cite{Abadi_Chu_Goodfellow_McMahan_Mironov_Talwar_Zhang_2016} is fed into an autoencoder during the training process. \cite{Abadi_Chu_Goodfellow_McMahan_Mironov_Talwar_Zhang_2016} uses differentially private stochastic gradient descent (DPSGD), where clipping operation is applied and Gaussian noise is added to the computed gradients. Then, the encoder is released and published, while the decoder is dropped. Following this, to encode and generate new differentially private data, a small amount of original data is fed into the encoder by the trainer. This new encoded/generated data is leveraged to train any machine learning models in the future which guarantees privacy. Thus, since it uses not only private data, but also public data, it provides high utility while preserving the data. Moreover, \textbf{Krishna et al. \cite{Krishna_Gupta_Dupuy_2021}} present an autoencoder-based differentially private text transformation (ADePT). ADePT is based on text-based autoencoder, such as LSTM sequence-to-sequence models, to transform the given text. It begins by transforming the given input text into a latent representation via the encoder (transformation phase). In this phase, a randomized algorithm is used. Specifically, the latent representation produced by the encoder is clipped, and then a Laplacian noise is added. Following this, it generates a new data based on the transformed data in previous phase through the decoder (generation phase). Thus, this model preserves the data privacy while maintains the dataset quality.

\paragraph{Generative Adversarial Network (GAN) \cite{Goodfellow_Pouget-Abadie_Mirza_Xu_Warde-Farley_Ozair_Courville_Bengio_2014}.} GAN is unsupervised learning that consists of two networks, namely, the generator (G) and the discriminator (D). The generator tries to fool the discriminator by synthesizing and generating samples that look like the original real inputs. In the meantime, the discriminator tries to distinguish between real and synthetic samples. Thus, these two networks contest with each other via a minimax two-player game. To protect the privacy of the training data from leaking and revealing sensitive information, differential privacy (DP) is adopted in the training process of GANs.
\\\textbf{PATE-GAN \cite{Jordon_Yoon_Schaar_2018}} is proposed to produce synthetic tabular data while maintaining the privacy. In this model, PATE method \cite{Papernot_Abadi_Erlingsson_Goodfellow_Talwar_2016} is adopted to guarantee differential privacy. PATE method \cite{Papernot_Abadi_Erlingsson_Goodfellow_Talwar_2016} divides the training dataset into k disjoint subsets, and then, k classifiers (in PATE-GAN K teacher discriminators) are trained individually on each subset. To produce a differentially private output during testing/classifying a new sample in PATE, a noisy aggregation of classifier outputs is performed. Furthermore, Jordon et al. propose to train a student discriminator with the synthesized data. The outputs of these synthesized data are given by the teachers through PATE method \cite{Papernot_Abadi_Erlingsson_Goodfellow_Talwar_2016}. Hence, the student network is trained privately. \textbf{DPGAN \cite{Xie_Lin_Wang_Wang_Zhou_2018}} is introduced to generate synthetic images without compromising privacy. It works by adding designated noise on the discriminator's gradients during the training process. In addition, gradient clipping is applied to enforce iterative gradient descent. \textbf{GANobfuscator \cite{Xu_Ren_Zhang_Zhang_Qin_Ren_2019}} is proposed to alleviate information leakage when GAN is used. In particular, a designated noise is added to the gradients during the learning process, and gradient clipping is leveraged to enhance the stability of the training process and improve the privacy as in DPGAN \cite{Xie_Lin_Wang_Wang_Zhou_2018}. However, in DPGAN \cite{Xie_Lin_Wang_Wang_Zhou_2018}, the model weights are clipped to a bounded box $[-c_{p},c_{p}]$ which in turn automatically bounds the gradients by constant $c_{g}$. Meanwhile, GANobfuscator \cite{Xu_Ren_Zhang_Zhang_Qin_Ren_2019} incorporates wasserstein GAN (WGAN) \cite{Gulrajani_Ahmed_Arjovsky_Dumoulin_Courville_2017} and adaptive clipping. In adaptive clipping, public data is used to adjust the gradients' clipping bounds throughout the training process.

\section{DP Future Research Directions}
\label{sec:others}
Although the aforementioned surveyed studies have shown great success in preventing machine learning models from leaking sensitive information and satisfying differential privacy requirement, there are some existing challenges that need to be addressed in future work. To this end, in this section, future research directions for differential privacy with machine learning algorithms are discussed.
\begin{itemize}
\item One major element that plays a key role in differential privacy is $\epsilon$ value since the amount of added noise depends on this value. The amount of added noise in turn affects the utility and accuracy of the model. The larger the $\epsilon$ value, the less noise is added; and hence, the model is more accurate but might be less privacy-preserving and vice versa. Thus, choosing the proper $\epsilon$ value is crucial; and consequently, more research should be conducted to determine the right $\epsilon$ value automatically.
\item Further, differential privacy is a hot topic among researchers and in academic settings. However, it is less used in industry and practical applications because its privacy guarantee is strong which might affect the utility and accuracy of the models. Thus, one possible direction for future work can be through leveraging a relaxation to DP, i.e., ($\epsilon$; $\delta$)-LDP, on industrial practical applications. Since some dimensions/variables are not sensitive and cannot reveal any sensitive information, a relaxed DP should be used.  For instance, country, state, or even city is not sensitive. Hence, participants probably do not care to reveal this information, but they really care about the precise and exact location which should have a strong and high level of privacy-preserving. 
\item Moreover, a hybrid method, that allows some participants to report their real values while perturbing other participants' values, may help in reducing the amount of noise added to the dataset. Thus, this might enhance the overall model accuracy while still protecting the data because privacy preferences are different among participants. Having an access even to a small subset of public dataset without perturbation may aid in estimating the hyper-parameters; and thus, improving the model accuracy and utility. 
\item In addition, an adjustable and adaptive privacy budget is a good approach instead of static and equal assignment of the privacy budget to every layer of the model.
\item Furthermore, using DP with complex machine learning models has many challenges. First, differentially private complex machine learning models increase the information leaking risk. Hence, they require excessive and more sanitization. This might lead to distort the data which in turn might affect the utility and accuracy of the models. Second, differentially private complex machine learning models are probably time-consuming in terms of computation. Therefore, it could be hard for researchers to incorporate DP with complex machine learning models. Consequently, new designated distributed protocols might be one possible solution for DP with complex models.
\item Another path for future work could be by incorporating cryptography techniques with differential private machine learning algorithms, particularly, through Distributed Differential Privacy (DDP).
\item Since differential privacy has its limitations especially when the dataset size is large due to time complexity and data utility, leveraging other privacy preserving approaches with machine learning should be explored. Examples of these other privacy preserving techniques are outline as follows.
\begin{itemize}
\item \textbf{Homomorphic encryption (HE) \cite{Acar_Aksu_Uluagac_Conti_2017}:} The data owner encrypts the data before sending the data to the model especially if the data provider uses the cloud to train the model. Then, the cloud homomorphically trains the model on the encrypted data and returns the encrypted result. Following this, the data owner can decrypt and share the result. Hence, the participants' data are protected; however, it is time-consuming where the complexity of computation is increased.
\item \textbf{Secure multiparty computation (MPC) \cite{Lindell_2020}:} This technique allows training the model on data from different parties without having the parties to share their data with any party. Thus, the data privacy is preserved; however, it is time-consuming even though it is less computation than HE.
\item \textbf{K-anonymity \cite{Bayardo_Agrawal_2005}:} Anonymization can be done by altering the data before feeding it to the model. In particular, k-anonymity modifies the data with k value, where k denotes the number of indistinguishable records. Generalization and suppression \cite{Samarati_Sweeney_1998} are some techniques that can be used to modify data and achieve anonymization. Generalization works by replacing the data with a broader/interval one, i.e., age of 50 can be replaced with [50,59]. Suppression works by removing sensitive data either removing the entire attribute values or some values of the attribute in some records. The downside of this mechanism is that it is vulnerable to background knowledge attack and homogeneity attack.
\item \textbf{L-diversity \cite{Machanavajjhala_Gehrke_Kifer_Venkitasubramaniam_2006}:} It is an extension of k-anonymity \cite{Bayardo_Agrawal_2005}, where it requires at least L well-represented values to be existed for sensitive attributes for every equivalence class. One main issue with L-diversity is that when the whole rows in a table are distributed into few equivalence classes, semantic closeness could lead to information revealing. Thus, it might be vulnerable to similarity attack.
\item \textbf{T-closeness \cite{Li_Li_Venkatasubramanian_2007}:} It is an extension of L-diversity \cite{Machanavajjhala_Gehrke_Kifer_Venkitasubramaniam_2006}, where equivalence class is a T-closeness if the distance between the distributions of sensitive feature in the class and the distribution of the feature in the entire table is less than or equal a specific threshold based on the Earth Mover's Distance\cite{Rubner_2000}. This approach guarantees data privacy, but the data distribution may not be appropriate each time when implementing T-closeness.
\item \textbf{Condensation approach \cite{Aggarwal_Yu_2004}:} it starts by constructing restricted clusters in the dataset. Then, it produces pseudo-data based on the statistics of the constructed clusters. The restrictions on the clusters depend on the cluster size that should be selected in a way to maintain k-anonymity \cite{Bayardo_Agrawal_2005}. This approach efficiently preserves the data privacy, but it might disclose some sensitive data because of the similarity in terms of values between some records in the generated/constructed data and original data.
\item \textbf{Data distribution approach \cite{Yan_Fu_Tan_Ng_2008,Deng_Jiang_Long_2020}:} It works by distributing the data across multiple sites in two ways (horizontal and vertical distributions). Horizontal distribution \cite{Yan_Fu_Tan_Ng_2008} works by partitioning the records of the dataset across several entities, each partition has same attributes. Meanwhile, vertical distribution \cite{Deng_Jiang_Long_2020} works by partitioning the attributes across many entities, each attribute holds all records.
\end{itemize}
\end{itemize}

\section{Conclusion}
\label{sec:others}
Differential privacy (DP) is a rigid definition of privacy based on injecting noise into the data. Different approaches can be used to inject a specific amount of noise into the data. The noise can be added to the inputs, the outputs, the ground truth labels, the objective functions, or even to the gradients to mitigate the privacy concern and preserve the data. Different mechanisms can be leveraged, such as Laplacian, Gaussian, exponential, and randomized response mechanisms, just to name a few.
\\Recently, after the advancement in machine learning and deep learning, more research has been conducted in this field due to its capability in solving real-world problems. However, vulnerabilities and revealing sensitive information may be a big concern with machine learning. Thus, DP can be incorporated with machine learning models to preserve the data from leaking and revealing sensitive information.
\\Therefore, this paper reviews various differentially private machine learning algorithms grouped into two categories: traditional machine learning models and deep learning models. Additionally, it concludes by discussing the future outlook and directions for differentially private with machine learning algorithms.

\section*{Acknowledgments}
The first author would like to thank Umm Al-Qura University, in Saudi Arabia, for the continuous support.

\bibliographystyle{unsrt}  
\bibliography{references} 
\end{document}